\pdfoutput=1

\def\year{2019}\relax
\documentclass[letterpaper]{article} 
\usepackage{aaai19}  
\usepackage{times}  
\usepackage{helvet}  
\usepackage{courier}  
\usepackage{url}  
\usepackage{graphicx}  
\usepackage{amsmath}
\usepackage{amssymb}
\usepackage{amsthm}
\usepackage{algorithm}
\usepackage{algorithmic}
\usepackage{wrapfig}
\usepackage{multirow}
\usepackage{arydshln}

\newcommand*{\affaddr}[1]{{\large #1}}
\newcommand*{\affmark}[1][*]{\textsuperscript{#1}}
\newcommand*{\email}[1]{#1}

\newcommand{\tabincell}[2]{\begin{tabular}{@{}#1@{}}{\arraycolsep=2pt
\def\arraystretch{1}\noindent$\begin{array}{lll}#2\end{array}$}\end{tabular}}

\newcommand{\citet}[1] {\citeauthor{#1}~\shortcite{#1}}

\begin{document}

\title{\textsc{FAnDa}: A Novel Approach to Perform Follow-up Query Analysis}

\author{%
Qian Liu\affmark[\textdagger]\thanks{Work done during an internship at Microsoft Research.}, Bei Chen\affmark[\S], Jian-Guang Lou\affmark[\S], Ge Jin\affmark[$\lozenge$]$^*$, Dongmei Zhang\affmark[\S]\\
\affaddr{\affmark[\textdagger]Beihang University, Beijing, China}\\
\affaddr{\affmark[\S]Microsoft Research, Beijing, China}\\
\affaddr{\affmark[$\lozenge$]Peking University, Beijing, China}\\
\email{qian.liu@buaa.edu.cn;~\{beichen, jlou, dongmeiz\}@microsoft.com;~elvisking@pku.edu.cn}\\
}

\maketitle
\begin{abstract}

Recent work on Natural Language Interfaces to Databases (NLIDB) has attracted considerable attention. NLIDB allow users to search databases using natural language instead of SQL-like query languages. While saving the users from having to learn query languages, multi-turn interaction with NLIDB usually involves multiple queries where contextual information is vital to understand the users' query intents. In this paper, we address a typical contextual understanding problem, termed as follow-up query analysis. In spite of its ubiquity, follow-up query analysis has not been well studied due to two primary obstacles: the multifarious nature of follow-up query scenarios and the lack of high-quality datasets. Our work summarizes typical follow-up query scenarios and provides a new FollowUp dataset with $1000$ query triples on $120$ tables. Moreover, we propose a novel approach \textsc{FAnDa}, which takes into account the structures of queries and employs a ranking model with weakly supervised max-margin learning. The experimental results on FollowUp demonstrate the superiority of \textsc{FAnDa} over multiple baselines across multiple metrics.     

\end{abstract}

\begin{table*}[t]
    \centering
    \scalebox{0.9}{
	\begin{tabular}{c|l}
    \hline
		\multicolumn{1}{c|}{\textbf{Scenario}} & \multicolumn{1}{c}{\textbf{Example}} \\
        \hline
        Analytics    & \tabincell{l}{
          \text{Precedent } &: \text{In 1995, is there any network named CBC?} \\ 
          \text{Follow-up }  &: \text{Any TSN?} \\ 
          \text{Fused }  &: \text{In 1995, is there any network named TSN?} 
        } \\
        \hline
        Compare   & \tabincell{l}{
          \text{Precedent }  &: \text{How much money has Smith earned?} \\ 
          \text{Follow-up }  &: \text{Compare it with Bill Collins.} \\ 
          \text{Fused }  &: \text{Compare money Smith earned with Bill Collins.}
        } \\
        \hline
        Calc \& Stats   & \tabincell{l}{
          \text{Precedent }  &: \text{List all universities founded before 1855.} \\ 
          \text{Follow-up }  &:\text{Show their number.} \\  
          \text{Fused }  &: \text{Show the number of all universities founded before 1855.} 
        } \\
        \hline
        Extremum    & \tabincell{l}{
          \text{Precedent }  &: \text{Which stadium has the most capacity?} \\ 
          \text{Follow-up }  &: \text{Which get the highest attendance?} \\  
          \text{Fused }  &: \text{Which stadium get the highest attendance?} 
        } \\
        \hline
        Filter    & \tabincell{l}{
          \text{Precedent }  &: \text{How many roles are from studio paramount?} \\ 
          \text{Follow-up }  &: \text{List all titles produced by that studio.} \\  
          \text{Fused }  &: \text{List all titles produced by studio paramount.} 
        } \\
        \hline
        Group    & \tabincell{l}{\text{Precedent }  &: \text{Show the industry which has the most companies?} \\ 
          \text{Follow-up }  &: \text{Show in different countries.} \\  
          \text{Fused }  &: \text{Show the industry which has the most companies in different countries.}
        } \\
        \hline
        Sort  & \tabincell{l}{\text{Precedent }  &:
          \text{Show all chassis produced after the year 1990.} \\ 
          \text{Follow-up }  &: \text{Sort them by year.} \\  
          \text{Fused }  &: \text{Show all chassis produced after the year 1990 and sort by year.} 
        } \\
        \hline
        Search   & \tabincell{l}{
          \text{Precedent }  &: \text{What position did Sid O'Neill play?} \\ 
          \text{Follow-up }  &: \text{Which players else are in the same position?} \\  
          \text{Fused }  &: \text{Which players play in the position of Sid O'Neill excluding Sid O'Neill?} 
        } \\
        \hline
	\end{tabular}
	}
	\caption{Typical follow-up scenarios.}\label{table_scenrio}
\end{table*}

\section{Introduction}

Natural Language Interfaces to Databases (NLIDB) relieve users from the burden of learning about the techniques behind the queries. They allow users to query databases using natural language utterances, which offers a better interactive experience compared to conventional approaches. By using semantic parsing techniques, utterances are automatically translated to executable forms (e.g. Structured Query Language or SQL) to retrieve answers from databases. The majority of the previous studies on NLIDB assumes that queries are context-independent and analyzes them separately. However, if we want to make NLIDB systems conform to users' mental models, it is vital to take contextual information into account. As users often pose new queries based on the past turns during a multi-turn interaction with the NLIDB system in a conversation. For example, given a query utterance, {\textit{``Show the sales in 2017.''}}, the user can simply say {\textit{``How about 2018?''}} instead of the complete query {\textit{``Show the sales in 2018.''}}. In fact, \citet{bertomeu2006contextual} point out that, in their Wizard-of-Oz experiment, up to 74.58\% queries follow immediately after the question they are related to. 

In this paper, we focus on immediate follow-up queries, and we formulate the follow-up query analysis problem here. Consider a context-independent question and a question immediately following it, respectively named as \textit{precedent query} and \textit{follow-up query}. Generally speaking, the follow-up query, like \textit{``How about 2018?''} in the above example, would be too ambiguous to be parsed into executable SQL by itself. Therefore, follow-up query analysis aims to generate a \textit{fused query}, which resolves the ambiguity of the follow-up query in the context of its precedent query. Compared to the follow-up query, the fused query reflects users' intent explicitly and facilitates better downstream parsing. In reality, there are various scenarios for follow-up queries, which can make the problem challenging. \citet{setlur2016eviza} introduce the scenarios of single queries in their realistic system, inspired by which, we summarize typical scenarios of follow-up queries in Table~\ref{table_scenrio}. For instance, the follow-up query \textit {``Compare it with Bill Collins.''} aims to perform a comparison, and \textit {``Show their number.''} belongs to the scenario of calculation and statistics.

Several attempts have been made to analyze follow-up queries in specific datasets. For example, in the air travel domain, the ATIS dataset collects user queries, including follow-up ones, and their corresponding SQL from a realistic flight planning system~\cite{dahl1994expanding}. Using this dataset, \citet{miller1996fully} employ a fully statistical model with semantic frames;~\citet{zettlemoyer2009learning} train a semantic parser using context-independent data and generate context-dependent logical forms; and \citet{suhr2018learning} present a relatively complex sequence-to-sequence model. While the ATIS dataset is realistic, it is limited to a particular domain. All these methods are specific to it and hard to transfer across datasets. More recently, the Sequential Question Answering (SQA) dataset is proposed along with a search-based method~\cite{iyyer2017search}. However, SQA focuses on relatively simple follow-up scenarios, where the answer to follow-up queries is always a subset of the answer to the precedent query. 

While some of the previous efforts somehow follow the idea of semantic parsing methods, typical analysis scenarios, such as compare, group and sort, are not covered in ATIS or SQA. The lack of public high-quality and richer datasets makes the problem even more challenging. Taking all the aforementioned limitations into account, we build a new dataset and present a well-designed method for natural language follow-up queries. Our major contributions are:

\begin{itemize}
\item We build a new dataset named FollowUp~\footnote{Available at https://github.com/SivilTaram/FollowUp},which contains $1000$ query triples on $120$ tables.
To the best of our knowledge, it is the first public dataset that contains various kinds of follow-up scenarios. 
\item We propose a novel approach, \textbf{F}ollow-up \textbf{AN}alysis for \textbf{DA}tabases (\textsc{FAnDa}), to interpret follow-up queries. \textsc{FAnDa} considers the structures of queries and employs a ranking model with weakly supervised learning. It is parser-independent and can transfer across domains.
\item We conduct experimental studies on the FollowUp dataset. Multiple baselines and metrics are utilized to demonstrate promising results of our model.
\end{itemize}

\section{Follow-up Query Dataset}\label{sec-dataset}

We create FollowUp dataset with the purpose of offering a high-quality dataset for research and evaluation. We utilize tables from WikiSQL dataset~\cite{zhong2017seq2sql}, as they are realistic extracted from the web. 
Tables with identical columns are joined, from which we randomly select $120$ tables with at least $8$ rows and $1$ numerical column. Data is collected by crowdsourcing of $8$ workers, using the format of the triple (precedent query, follow-up query, fused query). The collection is completed through two phases. Firstly, workers write context-independent precedent queries according to the tables. To avoid monotonous queries, we provide several generic prompts to workers such as \textit{``Require sort with an obvious order.''}, as ~\citet{pasupat2015compositional}. Secondly, given precedent queries, workers write follow-up queries and the equivalent fused queries. We fastidiously provide $10$ examples for each follow-up scenario, so that workers can imitate the examples to write different kinds of follow-up scenarios. 

FollowUp dataset contains $1000$ triples on $120$ tables with a vocabulary of size about $2000$. All the example triples in Table~\ref{table_scenrio} are from the proposed FollowUp dataset, which has great diversity in follow-up scenarios. Instead of SQL, we collect fused queries in natural language because SQL queries require workers equipped with more expertise. Natural language queries also allow methods for follow-up query analysis to be independent of the semantic parser. Furthermore, this kind of annotation format is embraced by several works on interactive question answering
~\cite{raghu2015statistical,kumar2016non,kumar2017incomplete}.

\begin{figure*}[t]
	\centering
	\includegraphics[width=0.9\linewidth]{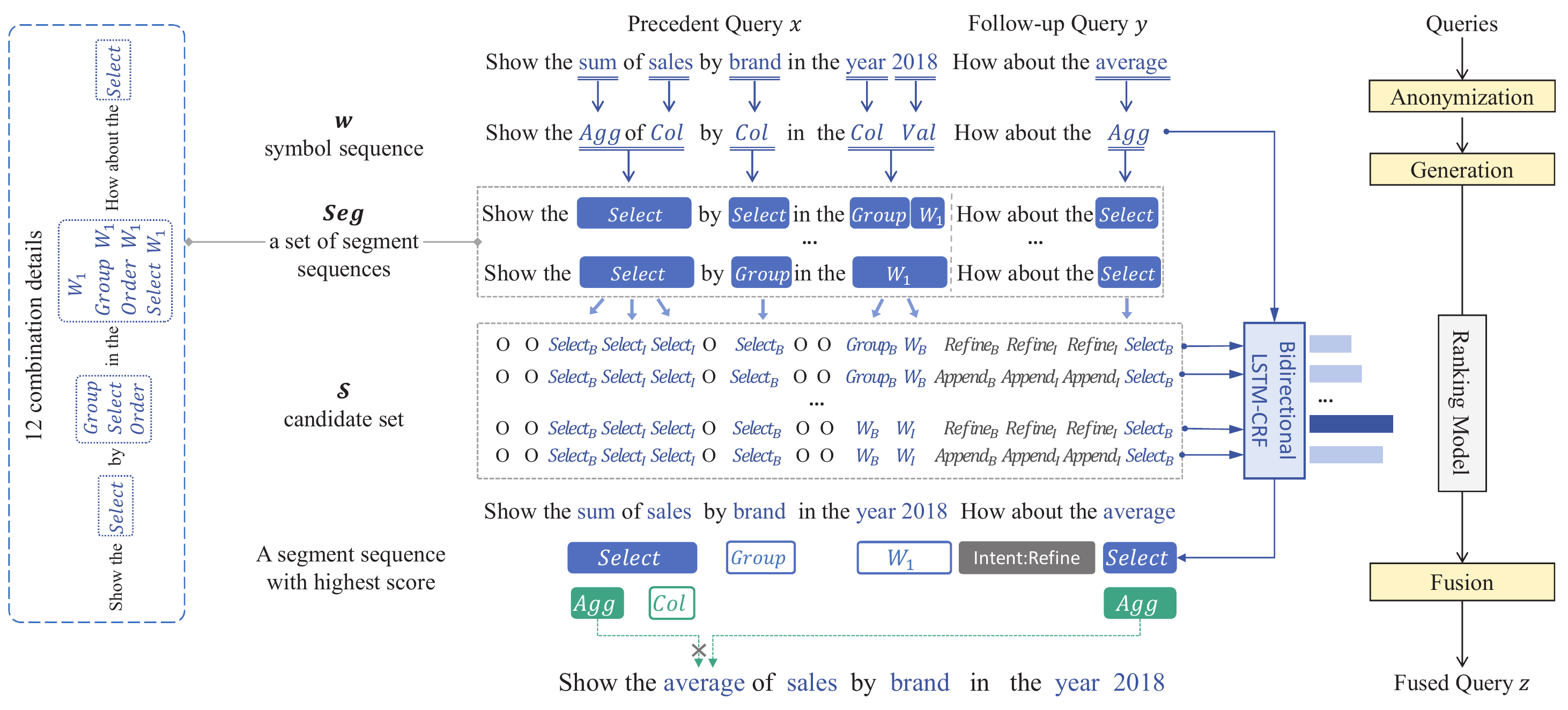}
	\caption{Illustration of \textsc{FAnDa}.}\label{fig_ranking_model}
\end{figure*}

\section{Follow-up Analysis for Database}\label{sec-fanda}

In this section, we present a novel approach \textsc{FAnDa}, by which the semantics of follow-up queries can be interpreted with precedent query. Taking the precedent query $x$ and the follow-up query $y$ as inputs, our goal is to obtain a complete fused query $z$. It has the same meaning with the follow-up query $y$ and can be processed by a downstream semantic parser all alone. Note that $x$, $y$ and $z$ are all natural language utterances. As the fused query $z$ always overlaps a great deal with the precedent and follow-up queries, it is natural to consider sequence-to-sequence based models. However, they are uninterpretable and require lots of training data, giving no consideration to the semantic structures of queries. In fact, $x$ and $y$ always have conflicting semantic structures. For example, given {\textit{``Show the sales in 2017.''}} and {\textit{``How about 2018 ?''}}, \textit{{``2018''}} conflicts with \textit{{``2017''}}, and only \textit{{``2018''}} should be kept. Therefore, in \textsc{FAnDa}, we carefully consider two-level structures: \textit{symbol}-level and \textit{segment}-level. Symbols are for words, and segments are for phrases related to SQL clauses. Three components, shown as yellow boxes in Figure~\ref{fig_ranking_model}, are devised to obtain the new fused query: (1) {\bf{Anonymization}}. Symbols are well-designed to simplify the queries, producing the symbol sequences. (2) {\bf{Generation}}. Segments are presented with compositional deduction rules, producing the best segment sequence. (3) {\bf{Fusion}}. Fusing $x$ and $y$ using the relationships among two-level structures, producing the fused query $z$.

\subsection{Anonymization}\label{sec_anon}

In query utterances, the words can be divided into two types: analysis-specific words and rhetorical words. Analysis-specific words indicate the parameters of SQL clauses explicitly, while rhetorical words form the sentence patterns. As shown in Figure~\ref{fig_ranking_model}, in the precedent query \textit{{``Show the sum of sales by brand in the year 2018''}}, the words \textit{{``sum''}}, \textit{{``sales''}}, \textit{{``brand''}}, \textit{{``year''}} and \textit{{``2018''}} are likely to be analysis-specific words, while the others are rhetorical words. As shown in Table~\ref{table_symbol}, we predefine $8$ types of \textit{symbol} for different analysis-specific words. Given a query, anonymization is to recognize all analysis-specific words in it, and replace them with the corresponding symbols to construct a symbol sequence. Following the example in Figure~\ref{fig_ranking_model}, the symbol sequence corresponding to $x$ should be {{``Show the $Agg$ of $Col$ by $Col$ in the $Col$ $Val$''}}. 

\begin{table}[t]
    \centering
		\begin{tabular}{c|c|c}
        \hline
			\textbf{Symbol} & \textbf{Meaning} & \textbf{Examples} \\
			\hline
			$Col$ & Column Name & sale, country \\
			$Val$ & Cell Value & 2018, Australia \\
			$Agg$ & Aggregation & sum, maximum, count \\
			$Com$ & Comparison & more, later, before \\
			$Dir$ & Order Direction & descending, ascending \\
			$Per$ & Personal Pronoun & it, he, them \\
			$Pos$ & Possessive Pronoun & its, his, their \\
			$Dem$ & Demonstrative & that, those, other \\
			\hline
		\end{tabular}
	\caption{Symbols for analysis-specific words.}\label{table_symbol}
\end{table}

The symbols $Col$ and $Val$ are table-related, while the others are language-related. For table-related symbols, the analysis-specific words can be found from the corresponding table of each query. Note that all the numbers and dates belong to $Val$. Replacing column names and cell values by $Col$ and $Val$ is the key to equip \textsc{FAnDa} with the ability to transfer across tables. For language-related symbols, \{$Per, Pos, Dem$\} are for pronouns, while the others are for different kinds of SQL operators. $Agg$ corresponds to aggregation function, $Com$ stands for comparison operator, and $Dir$ indicates the direction of ORDERYBY. The meanings of language-related symbols are limited to a narrow space, so it is viable to enumerate the most common analysis-specific words empirically. For example, $Pos \in \{$their, its, his, her$\}$ and $Agg \in \{$average, sum, count, maximum, $\cdots\}$. Both of the precedent query $x$ and the follow-up query $y$ are anonymized, and the resulting symbol sequences are denoted by $\hat{x}$ and $\hat{y}$ respectively. \footnote{If an analysis-specific word belongs to multiple symbols, several symbol sequences will be obtained. For example, ``those'' can be $Per$ or $Dem$. }

\subsection{Generation}

The symbol of an analysis-specific word reflects its intrinsic semantics, but ignores the content around it. Supposing we have parsed the query \textit{``Show the sum of sales by brand in the year 2018''} into a SQL statement. Although both \textit{``brand''} and \textit{``year''} are with the same symbol $Col$, they belong to different SQL clauses. Along with the adjacent $Val$ \textit{``2018''}, \textit{``year''} forms a clause WHERE year = 2018. As there are rhetorical words like \textit{``by''} around \textit{``brand''}, it forms a clause GROUPBY brand. Hence, inspired by SQL clauses, we design the structure \textit{segment} to combine symbols and capture the effect of rhetorical words. Each segment can be deduced by one or more adjacent~\footnote{Adjacent means that there is nothing but rhetorical words between two symbols and their distance in word level is less than a window size (4 in our experiments).} symbols according to the compositional deduction rule. Table~\ref{Table-compositional} shows the well-defined $8$ types of segments, along with their compositional deduction rules. $W$ and $P$ stand for $Where$ and $Pronoun$ respectively. Concatenating the symbol sequences $\hat{x}$ and $\hat{y}$ as a whole, the goal of generation is to obtain the correct segment sequence for it. However, there are multiple ways to combine symbols, making it problematic to acquire the correct segment sequence. Therefore, it is cast into a ranking problem. Firstly, symbols are combined to generate all possible segment sequences. Then, a ranking model is built to score these segment sequences and pick the best one as output.

\begin{table}[t]
	\centering
	\scalebox{0.85}{
	\begin{tabular}{c|l|c|l}
    \hline
		 \!\!\!\textbf{Segment}\! &\multicolumn{1}{c|}{\textbf{Rule}}&\!\textbf{Segment}\! &\multicolumn{1}{c}{\textbf{Rule}} \\
		\hline
        $Select$ &\!$[Agg + [Val]] + Col$ \!&$Group$ &\!\!$Col$ \\
		$Order$ &\!$[Dir] + Col $  \!& $P_1$ &\!\!$Per $   \\
        $W_1$ &\!$[{Col}] + [Com] + {Val}$ \!&$P_2$ &\!\!$Pos $  \\
        $W_2$ &\!$Col + {Com} + Col$ \!&$P_3$ &\!\!$Dem + Col$\!\!   \\
		\hline
	\end{tabular}}
	\caption{Segment types and compositional deduction rules. Square brackets indicate optional symbols.}\label{Table-compositional}
\end{table}

The compositional deduction rules originate from SQL clause syntax. For example, in Figure~\ref{fig_ranking_model}, ``\textit{sum of sales} ($Agg$ of $Col$)'' can make up a $Select$ segment, relevant to a SQL clause SELECT SUM(sales). There can also be multiple choices. ``\textit{year} ($Col$)'' can be $Select$, $Group$ and $Order$ alone, or be $W_1$ together with ``\textit{2018} ($Val$)''. To make the rules more robust, we leave out the order of symbols. For instance, both ([$Dir$], $Col$) and ($Col$, [$Dir$]) can be composed into segment $Order$. 

As the precedent query has a complete structure, the compositional deduction rules can be applied directly. However, ellipsis exists in the follow-up query, so all the symbols in the first $5$ rules become optional. Just as the follow-up case \textit{``How about the average''} in Figure~\ref{fig_ranking_model}, segment $Select$ can be deduced by a single $Agg$ without $Col$. Moreover, symbols in different queries cannot combine. Concatenating $\hat{x}$ and $\hat{y}$, we can generate multiple segment sequences and obtain the set $Seg$. For the examples in Figure~\ref{fig_ranking_model}, there are $12$ resulting segment sequences in $Seg$, as shown in the left blue dashed box. Then, a ranking model is built to pick the best segment sequence in $Seg$, which will be introduced in detail in Section~\ref{sec-model}. 

\subsection{Fusion} \label{subsec-fusion}

\begin{figure}[t]
	\centering
	\includegraphics[width=0.9\linewidth]{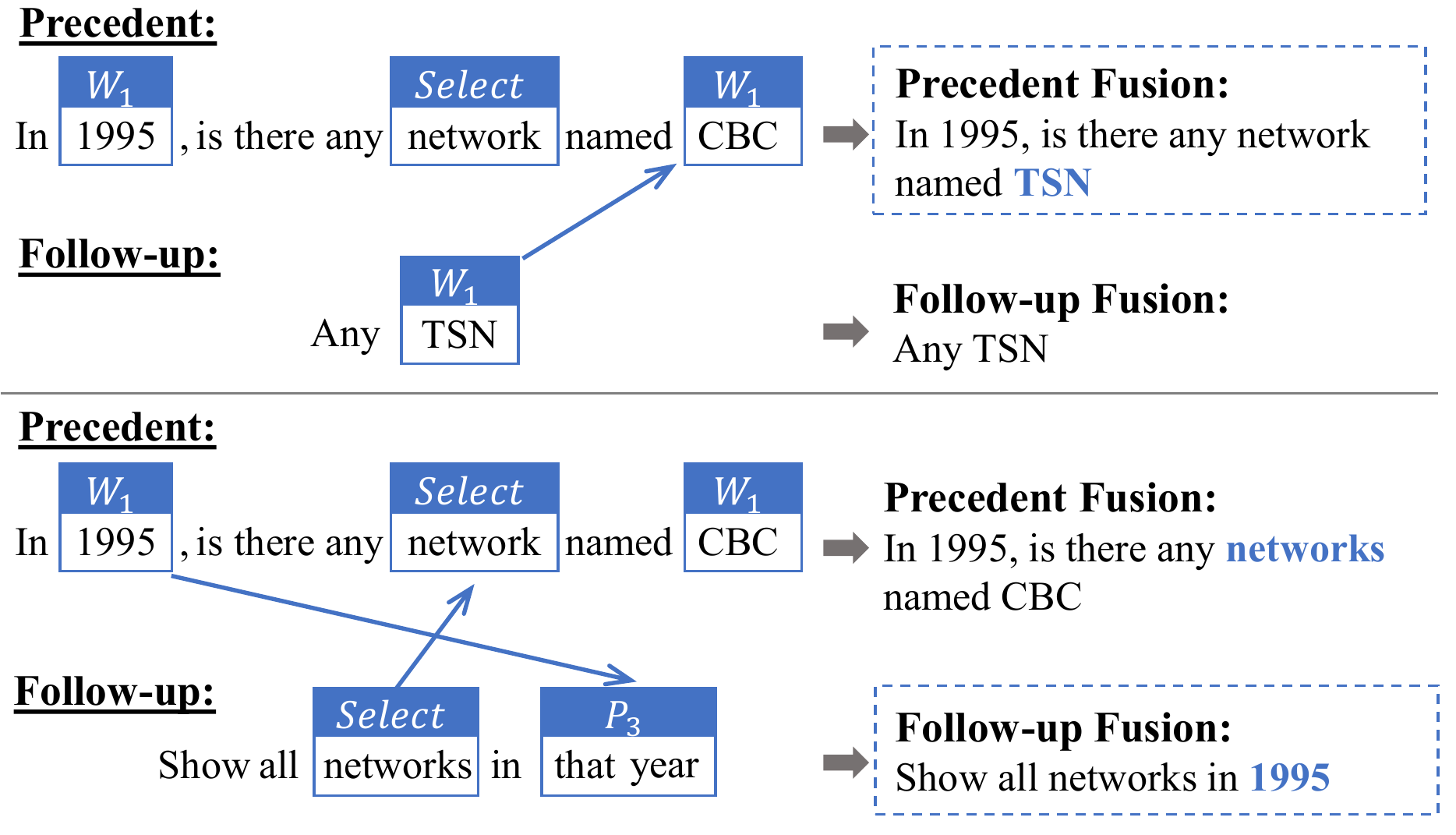}
	\caption{Two fusion cases.}\label{fig_fusion_example}
\end{figure}

Based on the symbol sequence and the best segment sequence, the fused query $z$ can be obtained. Breaking down the best segment sequence into two parts, one part corresponds to $x$, and the rest corresponds to $y$. There are two steps to accomplish the fusion. The first is to find conflicting segment pairs between the two parts. Conflicting means segments have the same or incompatible semantics. Generally speaking, segments of the same type conflict with each other. For instance, the two $Select$ segments conflict in the second case of Figure~\ref{fig_fusion_example}. However, there are particular cases. For $W_1$, segments conflict only if their inner symbols $Val$ are in the same column. It is the structured characteristic of tables that leads to incompatible semantics among these $W_1$ segments. As shown in the first case of Figure~\ref{fig_fusion_example}, instead of \textit{``1995''}, \textit{``TSN''} only conflicts with \textit{``CBC''}, for they are both in column \textit{``Network''}. For pronouns related segments, $P_{1}$, $P_{2}$ and $P_{3}$, we empirically design some semantic conflicting rules to resolve them, without considering ambiguities. For instance, $P_3$ ($Dem+Col$) conflicts with $W_1$ which describes the same column, such as \textit{``that year''} and \textit{``1995''} in Figure~\ref{fig_fusion_example}. $P_1$ ($Per$) conflicts with all words except those already in conflicting pairs, resulting in nested queries in $z$.

The second step is to perform fusion on these conflicting segment pairs. Generally, we fuse two segments by replacing one with the other. As indicated by different arrow directions in  Figure~\ref{fig_fusion_example}, we replace \textit{``CBC''} with \textit{``TSN''}, and replace \textit{``that year''} with \textit{``1995''}. When there is no pronoun, the replacement is symbol-level. Taking the example in Figure~\ref{fig_ranking_model}, ``\textit{sum of sales} ($Agg$ of $Col$)'' and ``\textit{average} ($Agg$)'' are both $Select$ segments and conflict with each other. Then only $Agg$ \textit{``sum''} is replaced by \textit{``average''}. Although replacement can be applied in most scenarios, it is not suitable for scenarios of compare, where the conflicting segments are presented for side-by-side comparison. Consider a query sequence \textit{``How much money has Smith earned? How about Bill Collins?''}, \textit{``Smith''} should be replaced by \textit{``Bill Collins''}. However, given \textit{``How much money has Smith earned? Compare with Bill Collins.''}, \textit{``Bill Collins''} should be added to \textit{``Smith''}. To distinguish the two different situations, we define two intents for follow-up queries: $Append$ for compare and $Refine$ for others. Thus, the output of ranking model turns into the best segment sequence and intent. There are various methods to perform intent classification, and we choose to regard them as two special segments. Finally, precedent fusion and follow-up fusion are obtained. We pick the follow-up fusion as output $z$ if it is different from the follow-up query. Otherwise, we choose the precedent fusion, as shown in the blue dotted boxes in Figure~\ref{fig_fusion_example}.

\section{Ranking Model} \label{sec-model}

As mentioned, in the process of generation, a ranking model is employed to pick the best segment sequence from $Seg$. In this section, we introduce how the ranking model works, followed by its learning process with weak supervision.

\subsubsection{Intent}

As previously stated, every follow-up query has an intent. We regard the two intent, $Refine$ and $Append$, as special segments. The intent of a follow-up query is related to its sentence pattern, which we believe contains all the rhetorical words before the first analysis-specific word. As in Figure~\ref{fig_ranking_model}, the sentence pattern \textit{``How about the''} is labeled as intent. Specifically, if there is no word before the first analysis-specific word, the intent is set as $Refine$.

\subsubsection{Mapping}
Inspired by named entity recognition~\cite{Tjong_Kim_Sang_2002}, we regard segment sequence ranking as the problem of tag sequence ranking. For simplicity, \{$W_1$, $W_2$\} are unified into $W$ and \{$P_1$, $P_2$, $P_3$\} are unified into $P$. An additional segment $O$, designed for $Others$, is employed for words without existing segments. 
Moreover, $O$ can also be deduced by the symbols \{$Per$, $Pos$, $Dem$\} in the situation where the pronouns are ambiguous, such as \textit{``that''} used as a conjunction.
Employing the IOB (Inside, Outside, Beginning) format~\cite{Ramshaw_1999}, we map $Seg$ into a set of tag sequences termed candidate set $\mathcal{S}$. 

One segment sequence usually maps to two tag sequences. As shown in Figure~\ref{fig_ranking_model}, the first two tag sequences are both from the first segment sequence, but have different intent tags $({Refine}_B, {Refine}_I, {Refine}_I)$ and $({Append}_B,{Append}_I,{Append}_I)$. The one with higher score represents the final intent of the follow-up query.

\subsubsection{Ranking}
Let $\boldsymbol{w}=(w_1, w_2,\cdots, w_{N})$ denote the concatenation of symbol sequences $\hat{x}$ and $\hat{y}$. The candidate set $\mathcal{S}=\{\boldsymbol{s}_1,\boldsymbol{s}_2,\cdots,\boldsymbol{s}_K\}$ contains tag sequence candidates, and the tag sequence can be denoted as $\boldsymbol{s}_{k}=(t_1^{k}, t_2^{k}, \cdots,t_{N}^{k})$. Our goal is to find the best candidate $\boldsymbol{s}^*$, that is:
\begin{equation}\label{eq_best}
\boldsymbol{s}^{*} = \mathop{\arg\max}_{\boldsymbol{s}\in \mathcal{S}} g(\boldsymbol{s} | \Theta ),
\end{equation}
where $g(\cdot | \Theta)$ is a score function given parameter set $\Theta$. To this end, we perform tag sequence candidates ranking using a bidirectional LSTM-CRF model~\cite{Huang2015BidirectionalLM} with weakly supervised max-margin learning. 
For each $w_i (i=1,\cdots,N)$, the model computes a hidden state $\mathbf{h}_i=[\mathop{\overrightarrow{\mathbf{h}}_i};\overleftarrow{\mathbf{h}}_i ]$, then the forward hidden state is:
\begin{equation}
\overrightarrow{\mathbf{h}}_i = {\overrightarrow{\mathbf{LSTM}}} \big( \phi(w_i);\overrightarrow{\mathbf{h}}_{i-1} \big),
\end{equation}
where $\phi$ is an embedding function initialized using Glove ~\cite{pennington2014glove}. Let $T$ denote the number of tags, and $\mathbf{f}_i$ denote the $T$-dimensional network score vector for $w_i$, which can be computed as:
\begin{equation}
\mathbf{f}_i = \mathbf{h}_i\mathbf{W},
\end{equation}
where $\mathbf{W}$ is the learned matrix. Let $\mathbf{A}$ denote the $T\times T$ transition matrix of CRF layer, and the entry $\mathbf{A}_{uv}$ is the probability of transferring from tag $u$ to $v$. Let $\theta$ denote the parameters of network in LSTM. 
Given $\Theta=\{\mathbf{A}, \mathbf{W}, \theta\}$, the score function for candidate $\boldsymbol{s}_k$ is defined as the sum of two parts: transition score by CRF and network score by bidirectional LSTM, which can be formulated as:
\begin{equation}\label{score-function-eq}
g(\boldsymbol{s}_k | \Theta ) = \mathop{\sum}_{i=1}^{N}(\mathbf{A}_{t^{k}_{i-1}t^{k}_{i}} + \mathbf{f}_i[t^{k}_{i}]),
\end{equation}
where $t^{k}_{i}$ is the corresponding tag of $w_i$ in candidate $\boldsymbol{s}_{k}$. 

\subsubsection{Weakly Supervised Learning}

Finally, we introduce how the bidirectional LSTM-CRF model is learned. As mentioned in Section~\ref{sec-dataset}, it is too expensive to annotate SQL, as well as tags. Hence, we utilize the gold fused query in natural language to learn, leading to the weak supervision.

For each tag sequence candidate $\boldsymbol{s}_k \in \mathcal{S}$, we can perform fusion based on its corresponding segment sequence and intent (Section~\ref{subsec-fusion}) , and obtain a natural language query $z_k$. Let $z^*$ denote the gold fused query. To compare $z_k$ and $z^*$, we process them by anonymization (Section~\ref{sec_anon}), while the pronouns are ignored. Then we check their symbols. If they have the same symbols with the same corresponding words, we call them \textit{symbol consistent} and put $\boldsymbol{s}_k$ in the positive set $\mathcal{P}$; otherwise, they are symbol inconsistent and $\boldsymbol{s}_k$ is put in the negative set $\mathcal{N}$. As we can see, $\mathcal{S}=\mathcal{P}\ {\cup}\ \mathcal{N}$. However, the tag sequences in $\mathcal{P}$ are not all correct. After fusion and anonymization, the sequences with wrong tags may result in symbol consistence by chance. Only one tag sequence in $\mathcal{S}$ may be correct, and the correct one is always in $\mathcal{P}$. As symbol consistence is the requirement of correctness on tags. Therefore, we calculate the scores of all tag sequences in $\mathcal{S}$, and select the highest ones from $\mathcal{P}$ and $\mathcal{N}$:
\begin{equation} 
\widehat{\boldsymbol{s}}_{p}=\mathop{\arg\max}_{\boldsymbol{s}\in \mathcal{P}}g(\boldsymbol{s}|\Theta), ~~~~~\widehat{\boldsymbol{s}}_{n}=\mathop{\arg\max}_{\boldsymbol{s}\in \mathcal{N}}g(\boldsymbol{s}|\Theta).
\end{equation}
Then a max-margin learning method is employed to encourage a margin of at least $\Delta$ between $\widehat{\boldsymbol{s}}_{p}$ and $\widehat{\boldsymbol{s}}_{n}$. Considering various lengths of different inputs, normalization factors are added to the scores. The hinge penalty is formulated as:
\begin{equation}
\max(0, {\Delta}-\frac{g(\widehat{\boldsymbol{s}}_{p}|\Theta)}{|\widehat{\boldsymbol{s}}_{p}|}+\frac{g(\widehat{\boldsymbol{s}}_{n}|\Theta)}{|\widehat{\boldsymbol{s}}_{n}|}),
\end{equation}
where $\Delta > 0$ is a hyperparameter. 

\section{Experiments}\label{sec-exper}

We evaluate our methods on the proposed FollowUp dataset, and split the $1000$ triples following the sizes $640$/$160$/$200$ in train/development/test. All the query utterances are pre-processed by anonymization (Section~\ref{sec_anon}). In the process of anonymization, dates and numbers are extracted for $Val$ using \textit{Spacy}\footnote{https://spacy.io/}, and person entities are recognized to handle personal pronouns. Moreover, for recognition of \textit{Col} and \textit{Val}, a simple matching algorithm is applied without considering synonyms.

We use three metrics to compare the two natural language queries: the output queries of our methods and the gold fused queries from the dataset.
(1) \textbf{Symbol Accuracy}. It is the proportion of the output queries that are symbol consistent (mentioned in Section~\ref{sec-model}) with the gold fused ones. It is used to measure the retention of critical information, but without considering the order of symbols. (2) \textbf{BLEU}. It automatically assigns a score to each output query based on how similar it is to the gold fused query~\cite{papineni2002bleu}. (3) \textbf{Execution Accuracy}. To further evaluate the validity of the output queries, we parse them into SQL and evaluate the execution accuracy manually. Specifically, 
we use \textsc{Coarse2Fine}~\cite{Lapata2018CoarsetoFineDF}, the state-of-the-art semantic parser on WikiSQL, to parse all the $200$ gold fused queries in the test set, and take the $103$ successful ones. Then the execution accuracy of the $103$ corresponding output queries is calculated. The other $97$ triples are excluded due to the incapability of the parser. 

\begin{table}[t]
    \centering
		\begin{tabular}{l|l|c|l}
        	\hline
			\multicolumn{2}{c|}{\textbf{Model}} & \multicolumn{1}{c}{\textbf{Symbol Acc }(\%)} & \multicolumn{1}{|c}{\textbf{BLEU }(\%)} \\		
			\hline
            \multirow{5}{*}{\rotatebox{90}{\textbf{Dev}}}
            &\textsc{Seq2Seq} &  ~0.63 $\pm$ {\small 0.00} &  21.34 $\pm$ {\small 1.14 } \\  
            &
            \textsc{CopyNet} &  17.50 $\pm$ {\small 0.87}~ & 43.36 $\pm$ {\small 0.54} \\ 
            &
            \textsc{S2S+Anon} &  18.75 $\pm$ {\small 0.95}~ & 41.22 $\pm$ {\small 0.33} \\ 
            &
            \textsc{Copy+Anon} &  25.50 $\pm$ {\small 2.47 } & 51.45 $\pm$ {\small 0.93} \\ 
            &
			\textsc{FAnDa} & \textbf{49.00} $\pm$ {\small 1.28 } &  \textbf{60.14} $\pm$ {\small 0.98} \\ 
            \hline
            \multirow{9}{*}{\rotatebox{90}{\textbf{Test}}}
            &
            \textsc{Concat} & 22.00 $\pm$ {\small --} ~~~~  & 52.02 $\pm$ {\small --}  \\
            &
            \textsc{E2ECR} & 27.00 $\pm$ {\small --} ~~~~  & 52.47 $\pm$ {\small --}  \\
            &
			\textsc{Seq2Seq} & ~0.50 $\pm$ {\small 0.22} & 20.72 $\pm$ {\small 1.31 } \\ 
            &
             \textsc{CopyNet} & 19.30 $\pm$ {\small 0.93 } &  43.34 $\pm$ {\small 0.45 } \\ 
            &
			\textsc{S2S+Anon} & 18.80 $\pm$ {\small 1.77 } & 38.90 $\pm$ {\small 2.45 } \\ 
            &
            \textsc{Copy+Anon} & 27.00 $\pm$ {\small 4.32 } & 49.43 $\pm$ {\small 1.11} \\ 
            &
			\textsc{FAnDa} & 47.80 $\pm$ {\small 1.14 }  & 59.02 $\pm$ {\small 0.54} \\ 
			&
            \quad-- Intent & 35.30 $\pm$ {\small 0.44 } &  55.01 $\pm$ {\small 0.86}\\ 
			&
            \quad-- Ranking & 24.30 $\pm$ {\small 6.70 } &  52.92 $\pm$ {\small 2.24 } \\ 
			&
            \quad+ Pretrain & \textbf{48.20} $\pm$ {\small 1.02 } & \textbf{59.87} $\pm$ {\small 0.43 }\\ 
            \hline
		\end{tabular}
	\caption{The results of symbol accuracy and BLEU scores.}\label{Table-results}
\end{table}

Our baselines include: 
(1) \textsc{\textbf{Concat}}: a simple method that directly concatenates precedent and follow-up queries;
(2) \textsc{\textbf{E2ECR}}: an end-to-end neural coreference resolution method. We perform evaluation using an already trained model provided by~\cite{Lee_2017} as it requires different training data;
(3) \textsc{\textbf{Seq2Seq}}: sequence-to-sequence model with attention~\cite{Bahdanau2014NeuralMT}; 
(4) \textsc{\textbf{CopyNet}}: sequence-to-sequence model with copying mechanism~\cite{Gu_2016}. For \textsc{Seq2Seq} and \textsc{CopyNet}, the input are the concatenation of the precedent and follow-up queries, and the features include word embeddings, part-of-speech tags, table information and so on;
(5) \textsc{\textbf{S2S+Anon}}: \textsc{Seq2Seq} with anonymized inputs;
(6) \textsc{\textbf{Copy+Anon}}: \textsc{CopyNet} with anonymized inputs. For \textsc{S2S+Anon} and \textsc{Copy+Anon}, inputs are anonymized using $Col$ and $Val$. For example, the utterance ``\textit{In 1995, is there any network named CBC? Any TSN?}'' is anonymized as ``\textit{In Val\#1, is there any Col\#1 named Val\#2? Any Val\#3?}''. 

\subsection{Follow-up Results}

\begin{figure}[t]
	\centering
	\includegraphics[width=1\linewidth]{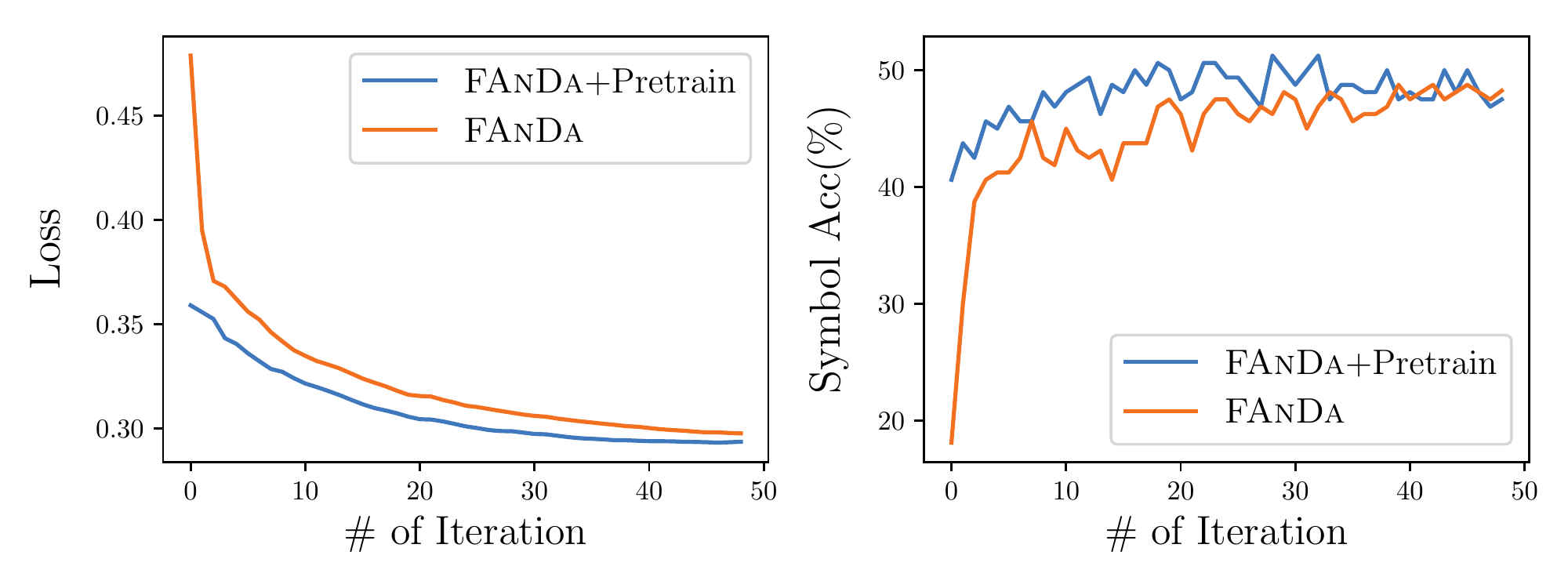}
	\caption{Convergence process on development set.}\label{fig_loss}
\end{figure}

Table~\ref{Table-results} shows symbol accuracies and BLEU scores on both development and test sets, where we run each experiment five times and report the averages. \textsc{FAnDa}--Intent means \textsc{FAnDa} equips all follow-up queries with intent $Refine$ without the intent classification; \textsc{FAnDa}--Ranking is to exclude the ranking model and select a segment sequence from $Seg$ randomly; and \textsc{FAnDa}+Pretrain is to annotate the tags of $100$ triples from train set and pre-train the bidirectional LSTM-CRF with supervised learning. We can observe that \textsc{FAnDa} significantly outperforms all the baselines, which demonstrates the effectiveness of our model. \textsc{S2S+Anon} and \textsc{Copy+Anon} get better results than \textsc{Seq2Seq} and \textsc{CopyNet} respectively, which demonstrates the importance of anonymization. The process of anonymization in \textsc{FAnDa} is well-designed and indispensable. Moreover, \textsc{FAnDa}--Intent performs worse than \textsc{FAnDa}, which shows the reasonability of distinguishing different intents. The bad performance of \textsc{FAnDa}--Ranking with a large standard deviation demonstrates the necessity of the ranking model. Unsurprisingly, \textsc{FAnDa}+Pretrain performs the best with the help of manual annotations. As shown in Figure~\ref{fig_loss}, pre-training with supervision can speed up the convergence, while the weakly supervised \textsc{FAnDa} has the competitive results.

\begin{table}[t]
    \centering
    \scalebox{0.85}{
		\begin{tabular}{l|c}
        	\hline
			\textbf{Model} & \textbf{Execution Accuracy} (\%) \\
            \hline
             \textsc{Concat} & 25.24\\
            \textsc{E2ECR} & 27.18 \\ 
             \textsc{Copy+Anon} &  40.77 \\ 
			\textsc{FAnDa} & \textbf{60.19} \\ 
			\hline
		\end{tabular}
		}
	\caption{The results of execution accuracies.}\label{Table-exec-acc}
\end{table}

Symbol accuracy is more convincing than BLEU, as the correctness of symbols is a prerequisite of correct execution. Table~\ref{Table-exec-acc} reports the execution accuracies on $103$ test triples. Due to the workload of checking the executable results manually, we only include baselines \textsc{Concat}, \textsc{E2ECR}, and the best baseline \textsc{Copy+Anon}. Results demonstrate the superiority of \textsc{FAnDa} over baselines in understanding context and interpreting the semantics of follow-up queries. 
It also shows that \textsc{FAnDa} is parser-independent and can be incorporated into any semantic parser to improve the context understanding ability cost-effectively.

\begin{table*}[t]
    \centering
    \scalebox{0.9}{
    \begin{tabular}{l|ll}
         \hline
         \textbf{No}&\multicolumn{2}{c}{\textbf{Case Analysis}} \\ 
         \hline
             \multirow{4}{*}{1}&
             \text{Precedent}\quad~~~~~: \text{What is the result, when the home team score is 2.4.6?} &
              \text{Follow-up}~: \text{What is the date?} \\ 
              &\multicolumn{2}{l}{\text{Gold Fusion}~~~\;\,: \text{What is the date, when the home team score is 2.4.6?} }\\
              &\multicolumn{2}{l}{\textsc{Copy+Anon}~~: \text{What is the date, the home team score is 2.4.6?}} \\
              &\multicolumn{2}{l}{\textsc{FAnDa}~~~~~~~~~~~\,: \text{What is the date, when the home team score is 2.4.6?} }\\
        \hline
        \multirow{4}{*}{2}&
              \text{Precedent}~~~~~~~~~: \text{Which is the draw number of Lowry?} &
              \text{Follow-up}~: \text{How about Laura?} \\ 
              &\multicolumn{2}{l}{\text{Gold Fusion}~~~\;\,: \text{Which is the draw number of Laura?} }\\
              &\multicolumn{2}{l}{\textsc{Copy+Anon}~~: \text{Which is the draw number of Lowry?}} \\
              &\multicolumn{2}{l}{\textsc{FAnDa }~~~~~~~~~~\,: \text{Which is the draw number of Laura?} }\\
        \hline
        \multirow{4}{*}{3}&
              \text{Precedent}~~~~~~~~~: \text{What are the names when elevation is feet?} ~~~~~~~~~~ &
              \text{Follow-up}~: \text{Of those, whose GNIS feature is 1417308?}\!\! \\ 
              &\multicolumn{2}{l}{\text{Gold Fusion}~~~\;\,: \text{Of names when elevation is feet, whose GNIS feature is 1417308?} }\\
              &\multicolumn{2}{l}{\textsc{Copy+Anon}~~: \text{What are the names when elevation is 1417308, whose GNIS feature is feet}?}\\
              &\multicolumn{2}{l}{\textsc{FAnDa}~~~~~~~~~~~\,: \text{What are the names when elevation is feet whose GNIS feature is 1417308?} }\\
         \hline
    \end{tabular}}
    \caption{Cases analysis of \textsc{Copy+Anon} and \textsc{FAnDa}.} \label{table_case}
\end{table*}

\subsection{Closer Analysis}

\begin{figure}[t]
	\centering
	\includegraphics[width=5cm]{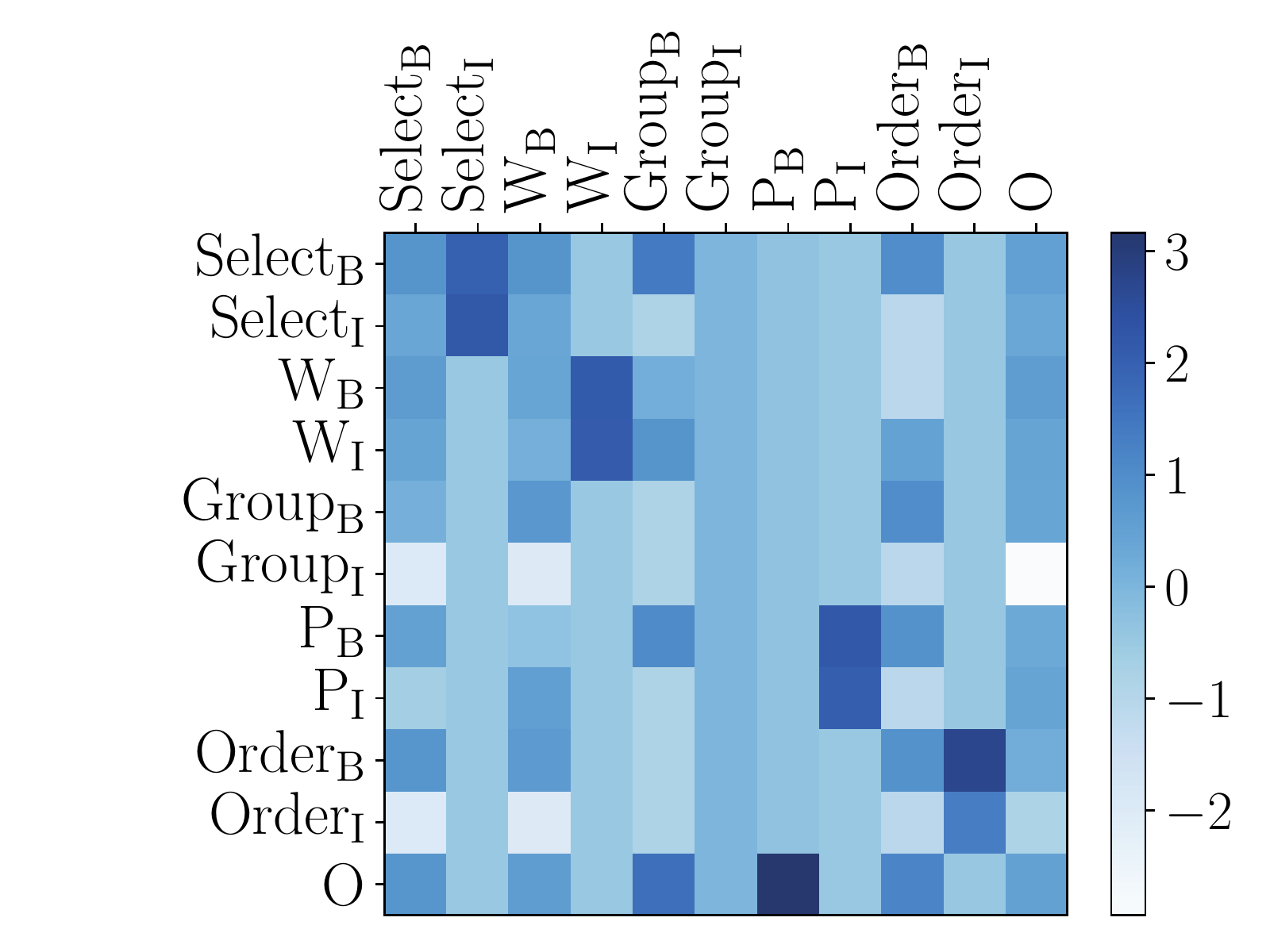}
	\caption{Transition matrix in CRF layer.}\label{fig_heat_map}
\end{figure}

\begin{figure}[t]
	\centering
	\includegraphics[width=0.9\linewidth]{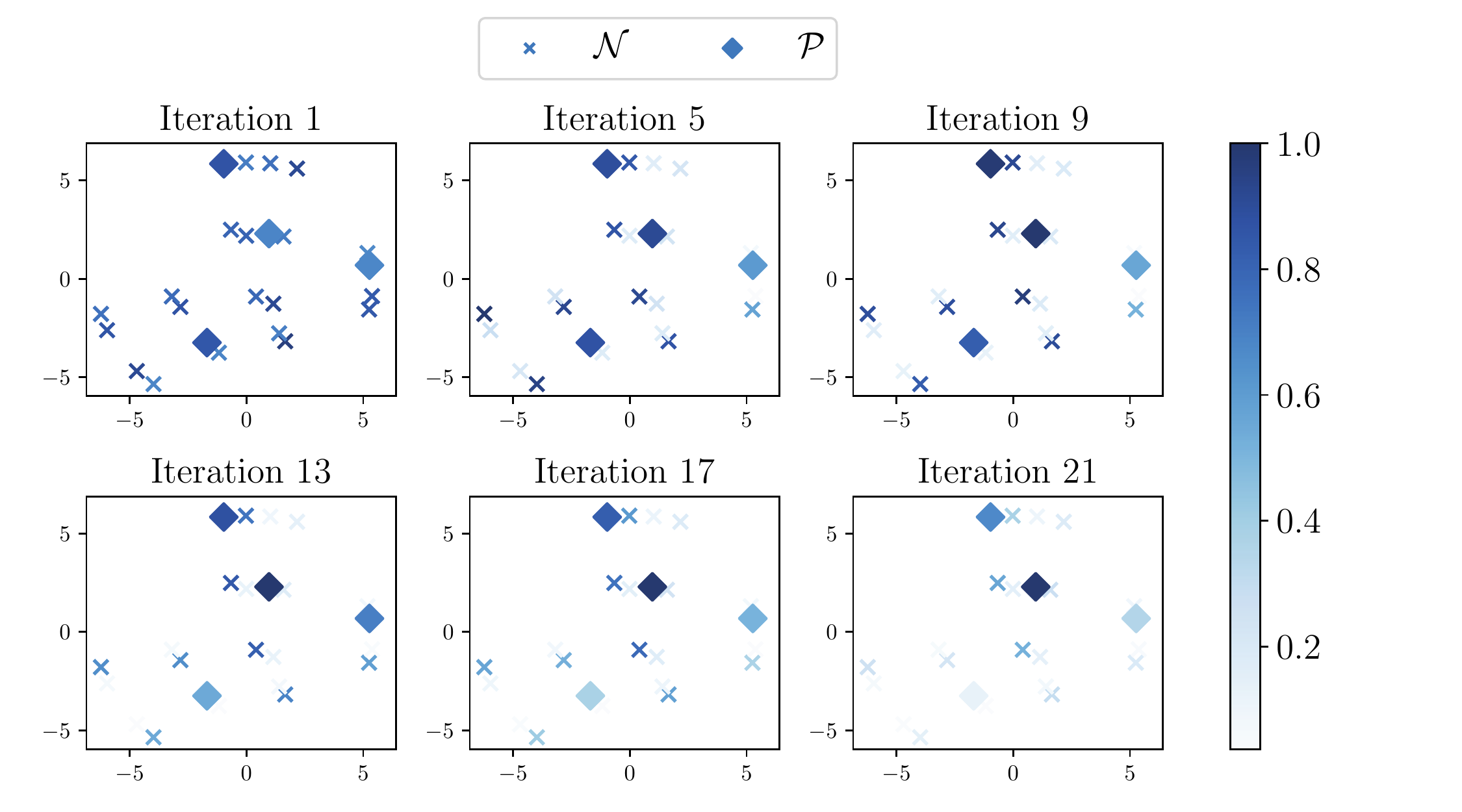}
	\caption{The evolution of candidate scores.}\label{fig_learning}
\end{figure}

Figure~\ref{fig_heat_map} shows the partial transition matrix $\textbf{A}$ of CRF layer in the ranking model. We observe that transition score from ${Tag}_{B}$ to ${Tag}_{I}$ is evidently higher than others, suggesting that CRF layer has learned which combinations of tags are more reasonable ($ Tag\in\{Select, W, Group, P, Order\}$). Figure~\ref{fig_learning} shows the evolution of candidate scores in $\mathcal{S}$ of a specific case, as training iteration progresses. In the coordinate system, each point represents a candidate. To scale the scores from different iterations into a unified space, we normalize them to range $[0,1]$. In Iteration 1, \textsc{FAnDa} assigns different but similar scores to all candidates in $\mathcal{P}$ and $\mathcal{N}$ with random initialization. From Iteration 5 to 21, the score distribution becomes increasingly skewed. The growing gap between the highest score and others verifies the effectiveness of max-margin learning. And from Iteration 13 to the end, the candidate with the highest score remains unchanged, indicating the stability of our weakly supervised learning.

Finally, we analyze three real cases in Table~\ref{table_case} and show the results of the generative model \textsc{Copy+Anon} and our model \textsc{FAnDa}. In case 1, both two models perform well. \textsc{Copy+Anon} puts \textit{``date''} in the position of \textit{``result''} according to the same context \textit{``what is the''}, indicating that generative models work well in the situation where a substantial overlap exists between precedent query and follow-up query. \textsc{FAnDa} can also deal with the situation, as bidirectional LSTM-CRF assigns \textit{``result''} and \textit{``date''} as $Select$ segment, and then \textit{``result''} is replaced with \textit{``date''}.

However, \textsc{Copy+Anon} performs worse than \textsc{FAnDa} mainly in two situations. The first situation is that there is no overlap. As in Case 2, \textsc{Copy+Anon} makes a mistake of ignoring \textit{``Laura''}, which should be used to replace \textit{``Lowry''}, indicating the weak reasoning ability of \textsc{Copy+Anon}. \textsc{Copy+Anon} only uses a learning-based approach, while \textsc{FAnDa} takes one step further by introducing the table structure to make judgments. The reason why \textsc{FAnDa} replaces \textit{``Lowry''} with \textit{``Laura''} is that they are both in column $Artist$. The second situation is that there is an ambiguous overlap. As in Case 3, there is a general word \textit{``is''} in front of both \textit{``feet''} and \textit{``1417308''}. Influenced by this, \textsc{Copy+Anon} confuses the positions of \textit{``feet''} and \textit{``1417308''}. \textsc{FAnDa} can solve the problem because it regards \textit{``elevation is feet''} and \textit{``GNIS feature is 1417308''} as separate segments.

\section{Related Work}\label{sec-work}

From the perspective of semantic parsing, our work is related to the analysis of context-independent queries, such as statistical parsing ~\cite{popescu2004modern,poon2013grounded} and sequence-to-sequence based methods ~\cite{jia2016data,iyer2017learning,Lapata2018CoarsetoFineDF}. Specifically,~\citet{palakurthi2015classification} utilize a CRF model to classify attributes into different SQL clauses, similar to our ranking model. From the perspective of follow-up analysis, there are multiple researches on context-sensitive conversation, such as open-domain response generation using neural networks~\cite{sordoni2015neural}, conversation agent using reinforcement learning~\cite{shah2018bootstrapping}, contextual question understanding for retrieval system~\cite{renconversational}, 
and non-sentential utterance resolution in question answering~\cite{kumar2016non,kumar2017incomplete}, which is similar to our baseline \textsc{S2S+Anon}.
Our work is also related to coreference resolution. The recent methods based on deep learning achieve the  state-of-the-art performances~\cite{long2016simpler,clark2016improving,Lee_2017}, from which we choose one as our baseline \textsc{E2ECR}. Moreover, several interactive visual analysis systems~\cite{setlur2016eviza,dhamdhere2017analyza,hoque2018applying} take context into account.

\section{Conclusion and Future Work}\label{sec-conclu}

For the purposes of research and evaluation, we create the FollowUp dataset that contains various follow-up scenarios. A novel approach, \textsc{FAnDa}, is presented for follow-up query analysis, which considers the structures of queries and employs a ranking model with weakly supervised max-margin learning. The experimental results demonstrate the effectiveness of our model. For future work, we are interested in extending our method to multi-turns and multi-tables. 

\section*{Acknowledgments}

We thank Yihong Chen, B\"{o}rje Karlsson, and the anonymous reviewers for their helpful comments.

\bibliography{follow-up}
\bibliographystyle{aaai}

\end{document}